# Optimal Transfer Learning Model for Binary Classification of Funduscopic Images Through Simple Heuristics

Principal Author: Rohit Jammula

Mentors: Vishnu Rajan Tejus, Shreya Shankar

# Abstract

Deep learning models have the capacity to fundamentally revolutionize medical imaging analysis, and they have particularly interesting applications in computer-aided diagnosis. We attempt to use deep learning neural networks to diagnose funduscopic images, visual representations of the interior of the eye. Recently, a few robust deep learning approaches have performed binary classification to infer the presence of a specific ocular disease, such as glaucoma or diabetic retinopathy. In an effort to broaden the applications of computer-aided ocular disease diagnosis, we propose a unifying model for disease classification: low-cost inference of a fundus image to determine whether it is healthy or diseased. To achieve this, we use transfer learning techniques, which retain the more overarching capabilities of a pre-trained base architecture but can adapt to another dataset. For comparisons, we then develop a custom heuristic equation and evaluation metric ranking system to determine the optimal base architecture and hyperparameters. The Xception base architecture, Adam optimizer, and mean squared error loss function perform best, achieving 90% accuracy, 94% sensitivity, and 86% specificity. For additional ease of use, we contain the model in a web interface whose file chooser can access the local filesystem, allowing for use on any internet-connected device: mobile, PC, or otherwise.

# Table of Contents

# 1 Introduction

## 1.1 Problem Statement

Eye diseases can often act stealthily, much like cancer; therefore, early diagnosis is of utmost importance. Many patients do not realize that they have an eye disease until they suffer irreversible damage to their vision. For instance, diabetic retinopathy afflicts a third of the world's 420 million diabetics. Moreover, there will soon be a critical shortage of medical workers and professionals. According to a 2016 study done by the Human Resources for Health, the global shortage for health care workers will become 15 million by 2030 (Liu et al., 2016). To most effectively fill the void, deep learning algorithms should be implemented on fundus eye exam (funduscopic) images.

## 1.2 Background

### 1.2.1 Funduscopic Image

Through the method of funduscopy, doctors use ophthalmoscopes to obtain a visual representation of the interior of the eye. The light rays from the ophthalmoscope pass through the pupil and illuminate the retina and the surrounding structures, known as the eyegrounds or fundus (Schneiderman, 1990). The lens on the ophthalmoscope magnifies the image, which represents the funduscopic image. These funduscopic (fundus) images provide the only noninvasive means to directly observe the retina, macula, optic disc, arteries, veins, and more. Fundus images can reveal both ocular diseases and indicators of systemic disease (Schneiderman, 1990).

### 1.2.2 Deep Learning

Deep learning, also known as representation learning, is a branch of machine learning that allows multiple "deep" processing layers to learn representations with multiple levels of abstraction (Bengio, 2012). It has been successful in tasks where large annotated datasets are widely available, for example object classification with supervised learning. Deep learning overcomes some of the limitations found in traditional machine learning models by automatically discovering the representations needed for a particular task, such as classification or regression, thus eliminating the need for a hand-engineered feature extractor. A deep learning model transforms representations of input data, starting with the raw input stream, into more abstract representations through a series of transformations.

### 1.2.3    Transfer Learning

Deep learning has improved medical image classification exponentially in the past few years, which means data acquisition and model complexity has skyrocketed (Raghu et al., 2019). However, this also means that producing a competitive deep learning model, especially from scratch, is becoming a less feasible option. Transfer learning allows one to import a base architecture designed to work on a completely different classification task, remove the top layers, replace them with more suitable ones, and alter the size of the input layer (Bengio, 2012). Most of the weights and biases intrinsic to the base architecture can be frozen or fine-tuned, which preserves their ability to discern more overarching elements of an image. But feeding in the new dataset will train the new unfrozen layers in the slightly altered architecture, which represents the more specific tasks of the classification. Sometimes, when the domain for the new dataset does not fit with the original domain, transfer learning models have difficulty generalizing to the new task (Kouw, 2018). Although transfer learning does not always demonstrate significant benefits over regular training methods (Raghu et al., 2019), it can be an effective tool to build on the success of a previous model.

### 1.2.4    Hyperparameter Demarcation

Hyperparameters can range from intrinsic factors such as architecture design, to surface-level factors such as batch size. As such, hyperparameters have fluid definitions. For our purposes, hyperparameters are defined as external factors more surface-level than model architecture and data augmentation (i.e. epoch number, batch size, optimizers, loss functions). They are easier to tune than more structural factors, such as data preparation and model architecture. Our "hyperparameters" are less significant: suited for granular control in later stages.

### 1.2.5    Additional Clarifications

Unless otherwise specified, we define the general concept of overfitting as train accuracy minus validation accuracy. Both metrics are recorded after the model has been fully trained. Sensitivity represents the ratio of true positives to all positives. Similarly, specificity is the ratio of true negatives to all negatives. A positive is defined as a member of the diseased class, and a negative is defined as a member of the healthy class.

## 1.3  Current Research

A few published deep learning algorithms already demonstrate the ability to perform classification on funduscopic images, to determine whether they have diseases, such as glaucoma, diabetic retinopathy, or macular degeneration. For instance, one such model has 95% and 81% accuracy for local and standard datasets of age-related macular degeneration (Langarizadeh et al., 2017). Another algorithm uses GoogLeNet, a preexisting architecture, to diagnose glaucoma and yields promising accuracy, despite the poor quality of some images in the dataset (Cerentinia et al., 2018). Diabetic retinopathy has received a particularly constant stream of attention from researchers who propose various types of deep learning models (Carson Lam et al., 2018).

## 1.4  Purpose

Although these existing models provide valuable information, they cannot replace doctors yet. So, a medical professional would still need to examine the eyes themselves and confirm the model's prediction, regardless of its complexity. Thus, there is a pressing need to create a general (healthy vs. diseased) algorithm, which has two major benefits. First, by definition, more data exists for a broad category (diseased eye) than any of its subcategories (glaucoma, for instance). For machine learning in general, it's commonly understood that more input data yields better generalization and less overfitting. Moreover, a layperson would not need to consult multiple, potentially conflicting models to determine whether they need to see a medical professional. They could simply determine whether their eyes are healthy or diseased and act accordingly; thus, the general model is more easily scalable. Alternatively, the general model could be used as an additional predictive tool to complement the diagnosis provided by a medical professional or more specific models. We also seek to leverage the model's scalability by building a web interface, accessible by anyone with an internet connection.

# 2 Development

Through most of the development process, we use Keras, a Python wrapper for the Tensorflow framework. This allows us to access to the vast majority of Tensorflow's features while maintaining ease of use and efficiency. Because deep learning is computationally expensive, we use a powerful P100 GPU to expedite the mathematical (especially matrix) calculations involved in training and evaluating the models. This was provided by the Jupyter kernels from Kaggle, which are free of charge and hold enough memory for dataset storage (13GB).

## 2.1 Data Acquisition

Different fundus image publishers use different imaging techniques. Therefore, it is crucial to ensure that all datasets have comparable numbers of both healthy and diseased fundus images. Moreover, no extraneous visual artifacts, other than indications of disease, can exist in any image. If they do, the extraneous artifacts must be common to both datasets, so that the model does not use them to differentiate the two classes. Consider a healthy vs. diseased model that uses two datasets: one with healthy images captured in low-light conditions, and another with diseased images captured in more intense light. The model might achieve high accuracy, but it would grossly overfit, because it only learns to differentiate between light and dark. Unfortunately, not many substantial datasets of fundus images are released publicly, due to patient confidentiality policies. Even fewer follow this essential prerequisite.

Using this information, we imported images from the EYEPACS diabetic retinopathy dataset (Gulshan et al., 2016) in a Kaggle competition and the ORIGA-650 glaucoma dataset (Zhang et al., 2010). The EYEPACS images have different severities of diabetic retinopathy, on an integer scale from 0-4. A rating of 0 means there is no diabetic retinopathy, and any other integer indicates that there is. We randomly sampled 3000 EYEPACS images from the healthy category, and 987 images from the diabetic retinopathy category. A very small number of images without diabetic retinopathy may have some other disease, but not enough exist to warrant full investigation. The ORIGA-650 dataset contains 482 healthy fundus images and 168 images with glaucoma (glaucomatous).

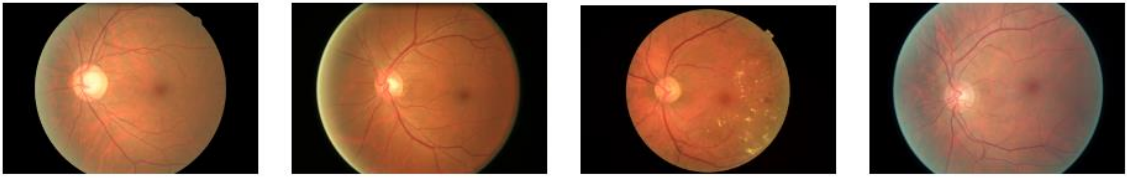

Figure 1: Examples from each major category listed respectively: Glaucoma ORIGA, Healthy ORIGA, Diabetic Retinopathy EYEPACS, Healthy EYEPACS (Gulshan et al., 2016; Zhang et al., 2010)

## 2.2  Data Preparation

The first step, to reduce color image size load is to resize every individual image into an array of size 128x128x3. There are 128 pixels per row, 128 per column, and three channels per pixel, each representing red, green, or blue. All values are integers ranging from 0 to 255 inclusive. This step specifically applies to supervised machine learning, because the model only accepts inputs with specific dimensions. After all, the algorithm is essentially a formula with a fixed number of input variables.

In order to balance out the dataset, we performed data augmentation on both healthy and glaucomatous categories of the ORIGA dataset. Data augmentation is a commonly used and essential practice, especially in the field of medical imaging where publicly released data is scarce. For each ORIGA image, we created b different orientations and zooms, with a background fill set to constant. We also created c different noise configurations, not including the control, in which each channel value was randomly shifted by an integer from -2 to 2 inclusive, but never escaped the bounds of 0 and 255. In total, an image can be replicated by a factor of b(c+1). For the 482 healthy ORIGA images, b=3 and c=1, and for the 168 glaucomatous ORIGA images, b=4 and c=3. This process yielded a total of 482*3*(1+1) = 2892 healthy ORIGA images, and 168*4*(3+1) = 2688 glaucomatous ORIGA images.

To form the final healthy dataset, we combined the 2892 healthy ORIGA images with the 3000 randomly sampled healthy EYEPACS images to get 5892 total healthy images. To form the final diseased dataset, we created 2 additional copies of each of the 987 diabetic retinopathy EYEPACS images (1974 copies + 987 originals = 2961 total EYEPACS diabetic retinopathy images); we then combined them with the 2688 glaucomatous ORIGA images to get a total of 5649 diseased images. In total, we had 11541 images to work with.

I then randomly shuffled them, and approximately allocated 60% (6924 images) to the training set, 20% (2308 images) to the validation set, and 20% (2309 images) to the test set.

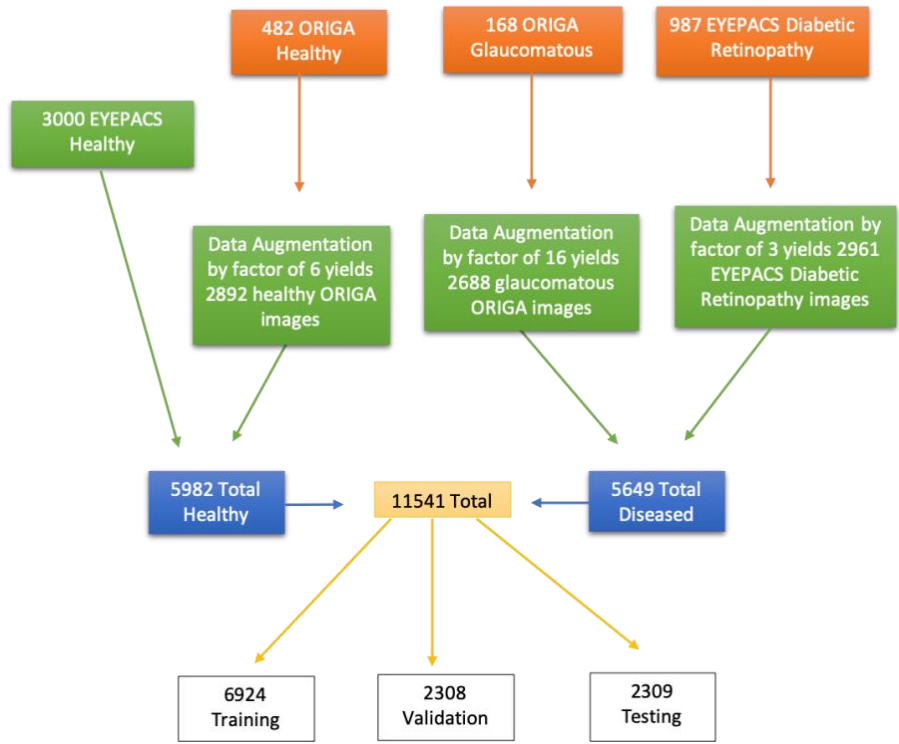

Figure 2: Flowchart Representation of Image Sorting

## 2.3  Baseline Model

Inception models are one of the most powerful convolutional neural networks (CNNs) trained for image data. Xception models take Inception's crucial depth-wise convolution to the extreme. Thus, we assumed that using Xception as a base architecture would act as a sufficiently robust baseline for our dataset. Moreover, transfer learning does not necessarily perform better than a regularly trained counterpart, as mentioned in 1.2.3. However, when repurposed with randomly initialized weights, the model performed poorly. Although the model hardly overfits, indicating proper generalization, it cannot compensate for the other metrics that indicate poor performance.

| Valid Accuracy | Overfitting | Valid Loss | Sensitivity | Specificity |
|---|---|---|---|---|
| 0.5841 | 0.0223 | 0.6948 | 0.4996 | 0.6699 |

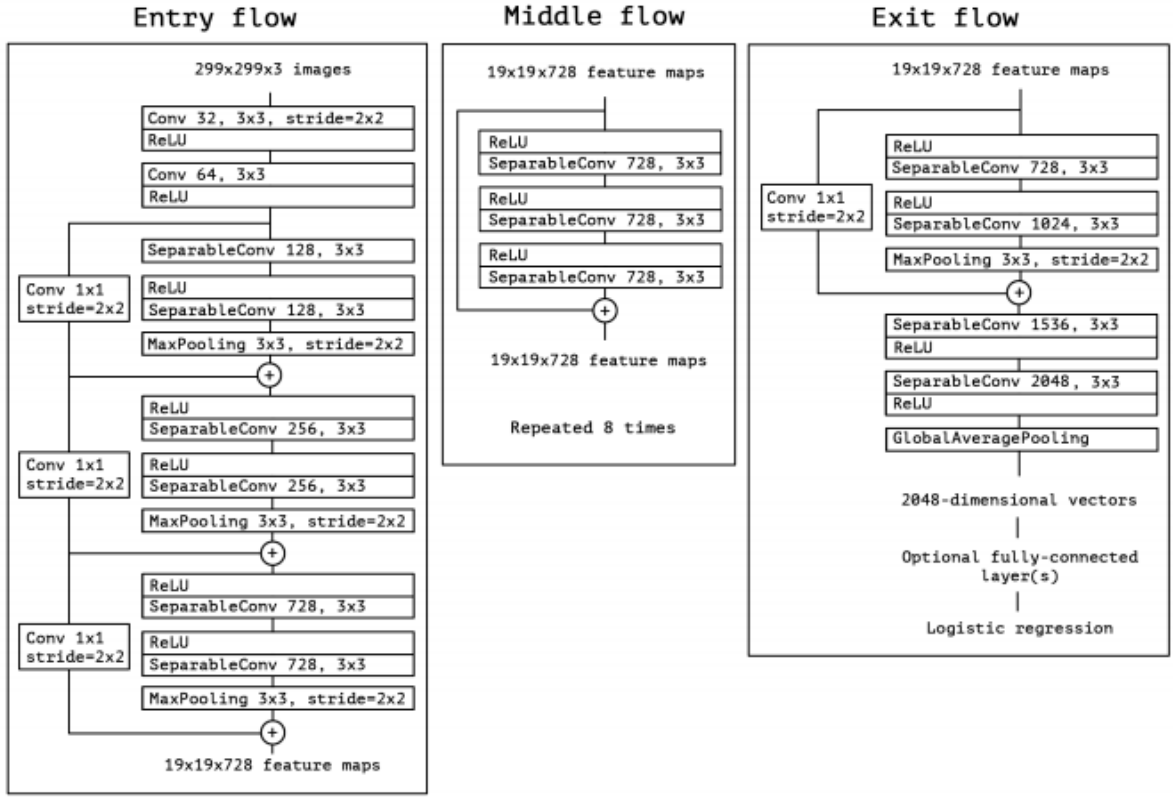

Figure 3: Convolutional flow of Xception model architecture, illustrating depthwise convolutions, making it ideal for image recognition. (Chollet, 2016)

## 2.4 Experiments

Pruning our approach to find the most effective base architecture before tuning hyperparameters, as defined in 1.2.4, is preferable to testing every possible combination of hyperparameters and base architectures. Many different types of hyperparameters can be tuned, including number of epochs, batch size, and learning rate. These adjustments generally alter the models' performance in an unpredictable manner. During preliminary testing on our dataset, hyperparameter tuning produced relatively mild shifts in performance. So, if a particular base architecture performs significantly better than the rest on our dataset, experimenting with different hyperparameters with a candidate base architecture would likely not make it a competitor to the optimal model.

Thus, performing hyperparameter optimization on only one base architecture saves time by reducing the number of experiments performed.

As such, we split the experimentation into the following two stages. Stage 1 determines which base architecture to start with, and Stage 2 determines which hyperparameters to set with the selected base architecture.

### 2.4.1   Default Settings

Nonetheless, some initial default conditions are always configured, no matter the model under evaluation. The top layers are always removed, and are replaced with a flatten command on the most recent layer, a dropout of 0.5, and the final Dense layer of size 2 with SoftMax activation, which outputs a probability vector determining the predicted label. The dimensions of the input layer are altered to match the dimensions of each resized image: 128x128x3. The batch size is 32 and there are 15 epochs. Finally, the batch normalization layers in the network, if they appear, must be unfrozen. This step ensures that their regularization properties can be properly utilized for the fundus image data we are feeding, instead of their original data, part of the ImageNet corpus for object recognition.

### 2.4.2   Evaluation Metric Ranking System

In a normal training pipeline, no definitive method exists for selecting the correct base architecture or hyperparameters, so we have created our own heuristic. In order of decreasing importance, we use these evaluation metrics: overfitting as defined in 1.2.5, validation accuracy, validation loss, sensitivity, and specificity. Although this process has not been formally published, we believe it has some utility. We've considered incorporating the F1 score as well, but decided against it for simplicity's sake.

After aggregating all the data, every model is ranked against one another, for each of the five evaluation metrics mentioned. We use ranks to mitigate the variations in raw metric values, especially the loss which has no defined range of values. Given N different models in a Stage, the highest value is always given a rank of 1, and the lowest value is always given a rank of N. The ranks in Stage 1 have no relation to the ranks in Stage 2. An individual model's overall score will be defined by this equation:

**OVERALL SCORE** = 3 • (overfit rank) + 2 • (N + 1 – accuracy rank) + 1.5 • (loss rank) + 1 • (N + 1 – sensitivity rank) + 0.25 • (N + 1 – specificity rank)

The highest overall score determines the model with best performance in each Stage. In the event of a tie, the model with the least number of total parameters would get a higher rank. Simpler models are more portable in the real world, so we want to encourage them.

We invoke the principle of mathematical plausibility to explain the subtraction from N, for the overfit and loss ranks. Because accuracy, sensitivity, and specificity are desirable, we wish to maximize them. Thus, the models with "lower" ranks (closer to 1), for these metrics, receive a greater overall score. On the other hand, models with "higher" ranks (closer to N), for undesirable metrics like overfitting and loss, still receive a greater overall score because they are not subtracted from N+1. The "+1" simply standardizes the overall scores and mostly does not influence rank. This way, the coefficients can be applied properly to each term. Without the "+1", the overfit term, for instance, could end up slightly de-prioritized when compared to the validation accuracy term.

Validation accuracy, and confusion matrices (essentially containers for sensitivity and specificity) are used in a wide variety of literature (Toghi & Grover, 2018). Moreover, sensitivity is more important than specificity, because positive is defined as the diseased class. Considering that this is a medical application, we must catch as many positive instances as possible.

### 2.4.3 Stage 1: Base architecture Selection Process, N=17

We use previously released classification models that have trained on several million ImageNet images. The seventeen transfer learning candidates for our base architecture are Xception, ResNet50, ResNet50V2, ResNet101, ResNet101V2, ResNet152, ResNet152V2, VGG16, VGG19, InceptionV3, InceptionResNetV2, MobileNet, DenseNet121, DenseNet169, DenseNet201, NASNetLarge, NASNetMobile. The default optimizer is rmsprop, and the default loss function is categorical cross-entropy.

**2.4.4   Stage 2: Hyperparameter Optimization, N=9**

When the chosen base architecture is selected, it must undergo a second stage of testing: hyperparameter tuning as described in section 2.4. In our case, however, these parameters produce granular variations. Moreover, our Keras wrapper randomly initializes the starting point for each instance of training. This randomness coupled with the relative insignificance of those variations indicates that the rankings for Stage 2 would effectively become a random number generator in a list. To avoid that scenario, we use hyperparameters that have greater impacts on the calculations themselves: for instance, optimizers and loss functions. For this experiment, we compare nine hyperparameter configurations, using 3 different optimizers and 3 different loss functions.

The optimizers will be rmsprop (RMS), Adam, and Adagrad. All of them show great promise in the development of neural networks. The RMS and Adam optimizers have been revolutionary in deep learning and works quite well in practice. Adagrad has similar proficiency with deep learning applications. The loss functions we choose to test are categorical cross-entropy (CCE), mean squared error (MSE), and mean absolute error (MAE). CCE is often considered as a standard loss function; we use it as a default throughout the testing for Stage 1. For many computer vision tasks such as scene understanding and activity recognition, L1 and L2 loss, the sum variations of MAE and MSE, respectively, are widely justifiable functions (Janocha & Czarnecki, 2017).

# 3   Results

## 3.1   Data Tables

Table 1: Stage 1

| Model | Overfitting | Validation acc. | Loss | Sensitivity | Specificity | Rank |
|---|---|---|---|---|---|---|
| **Xception** | **0.0952** | **0.9008** | **0.3468** | **0.9407** | **0.8603** | **1** |
| Resnet50 | 0.0914 | 0.8925 | 1.4468 | 0.9613 | 0.8227 | 2 |
| Resnet50V2 | 0.1296 | 0.8219 | 0.792 | 0.9355 | 0.7066 | 16 |
| Resnet101 | 0.0968 | 0.8921 | 1.1475 | 0.9355 | 0.848 | 6 |
| Resnet101V2 | 0.1165 | 0.8618 | 0.7693 | 0.9226 | 0.8 | 10 |
| Resnet152 | 0.0892 | 0.8938 | 1.5941 | 0.8942 | 0.8934 | 3 |
| Resnet152V2 | 0.1064 | 0.8826 | 0.4262 | 0.9011 | 0.8638 | 7 |
| VGG16 | 0.0699 | 0.7552 | 2.2893 | 0.7498 | 0.7607 | 8 |
| VGG19 | 0.0751 | 0.7171 | 2.5212 | 0.5701 | 0.8664 | 13 |
| InceptionV3 | 0.0890 | 0.7773 | 0.457 | 0.8426 | 0.7109 | 5 |
| InceptionResNetV2 | 0.1004 | 0.7409 | 0.6184 | 0.6578 | 0.8253 | 17 |
| MobileNet | 0.1202 | 0.8228 | 2.1897 | 0.9733 | 0.6699 | 15 |
| DenseNet121 | 0.0776 | 0.8193 | 0.7664 | 0.6939 | 0.9467 | 4 |
| DenseNet169 | 0.0438 | 0.6551 | 2.2782 | 0.3336 | 0.9817 | 11 |
| DenseNet201 | 0.0323 | 0.6937 | 4.342 | 0.7455 | 0.641 | 9 |
| NASNetLarge | 0.0363 | 0.6308 | 3.2208 | 0.6784 | 0.5825 | 14 |
| NASNetMobile | 0.0131 | 0.5789 | 2.9431 | 0.2614 | 0.9013 | 12 |

Table 2: Stage 2 - Xception Base Architecture

| Model | Overfitting | Val accuracy | Loss | Sensitivity | Specificity | Rank |
|---|---|---|---|---|---|---|
| RMS, CCE | 0.1395 | 0.8098 | 0.6115 | 0.9587 | 0.6588 | 9 |
| RMS, MSE | 0.1118 | 0.8544 | 0.1195 | 0.8521 | 0.8568 | 3 |
| RMS, MAE | 0.0890 | 0.8011 | 0.2032 | 0.7325 | 0.8707 | 7 |
| Adam, CCE | 0.1097 | 0.8817 | 0.3780 | 0.9251 | 0.8399 | 4 |
| **Adam, MSE** | **0.0846** | **0.9015** | **0.0925** | **0.9413** | **0.8560** | **1** |
| Adam, MAE | 0.0783 | 0.8219 | 0.1827 | 0.8667 | 0.7222 | 2 |
| Adagrad, CCE | 0.1040 | 0.8528 | 0.3540 | 0.8418 | 0.8629 | 5 |
| Adagrad, MSE | 0.1158 | 0.8133 | 0.1416 | 0.9036 | 0.7213 | 8 |
| Adagrad, MAE | 0.0787 | 0.7595 | 0.2440 | 0.7825 | 0.7362 | 6 |

## 3.2   Test Set Verification

The model with the best results uses the Xception base structure with an Adam optimizer and the mean squared loss function. To verify the validation results in the fifth row of Stage 2, we compare them to the test results. We then note the absolute value of the differences between the validation and testing versions of each metric. They represent their own versions of overfitting separate from the one defined in 1.2.5. The absolute values of the differences (metrics of overfitting) are extremely small, indicating that the model as a whole generalizes to the test set as well as it does with the validation set.

| Test Acc | Test Sensitiv | Test Spec | Val Acc | Val Sensitiv | Val Spec |
|---|---|---|---|---|---|
| 0.8748 | 0.8952 | 0.8561 | 0.9015 | 0.9413 | 0.8560 |

| |Accuracy Difference| | |Sensitivity Difference| | |Specificity Difference| |
|---|---|---|
| 0.0267 | 0.0461 | 0.0001 |

# 4   Implementation

## 4.1   Web App Creation

To broaden access to this tool, we decided to create a web-based application, which can be accessed by anyone with an Internet connection. We use the Flask framework to directly interface between the Python backend, used to prepare images and run the model, and the HTML/CSS frontend, which displays and updates the client side. We use a computer's localhost server to test run and develop the app.

When preparing the user input, we employ the same strategies we used in training. We first take the image and resize it to 128 pixels x 128 pixels x 3 color channels. We make sure only to take the first 3 channels, to account for the fourth channel present in PNG images. Then, we take the packaged model and use a Keras method to run the example. To relay the prediction output and reload the page with the result, we use GET and POST requests and dynamically update the user interface.

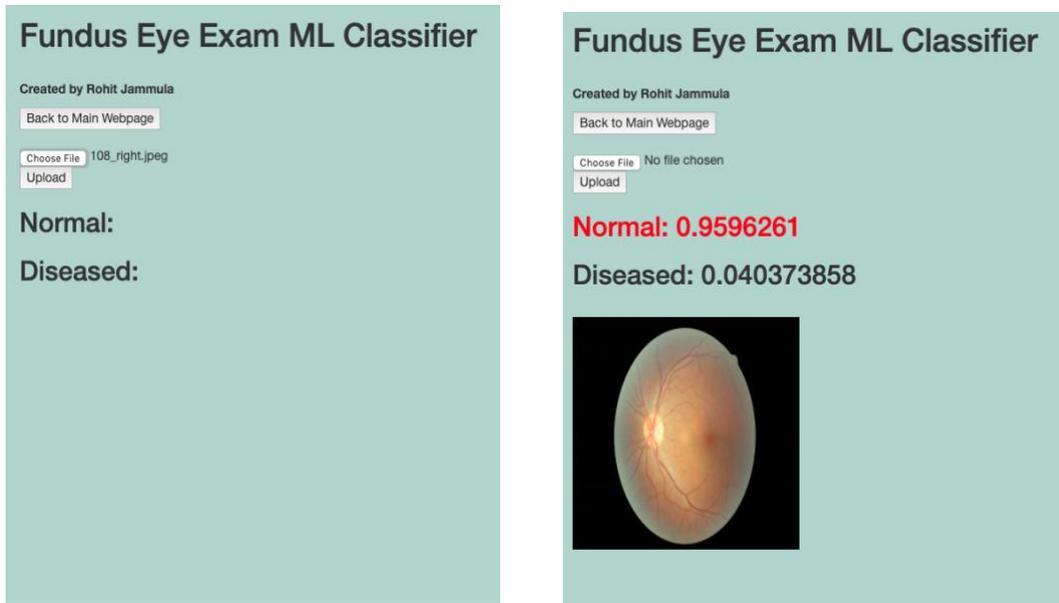

Figure 4: The image on the left displays the name of the chosen file. The image on the right displays the fundus eye exam, and highlights the class that corresponds with the prediction given by the model.

## 4.2 Deploying and Scaling

A major contributor to the success and adoption of deep neural networks has been the increase in available computational power and recent advances in parallel computing, such as CUDA by NVIDIA that runs on Graphics Processing Units (GPUs).

As more users access our web app and upload funduscopic images, our implementation needs to have mechanisms that can allocate resources and scale up to demand accordingly, as well as scale down when there is less usage. For creating, scheduling, and updating jobs, load balancing, as well as ensuring more robustness in a situation where a compute node might go down, we host the AI-Sight web app on Google Cloud Platform's container-orchestration system, also known as Kubernetes.

# 5    Discussion and Implications

Given this particular two-stage approach, novel heuristic, and data setup, we find that the optimal transfer learning model for binary classification on funduscopic images is the Xception base architecture with an Adam optimizer and a mean-squared loss function. In Section 3.2, we verified this with the generalization results from the test set, as well as the validation set.

## 5.1    Comparison with Baseline

Our final model performs substantially better than the baseline with randomly initialized weights, whose data is referenced in 2.3. We cannot confirm this by comparing overall score results, because the formula depends on ranks which would be infeasible for only two models. So, we must compare both models through each evaluation metric individually.

| Model | Overfitting | Validation Accuracy | Loss | Sensitivity | Specificity |
|---|---|---|---|---|---|
| **Baseline** | 0.0223 | 0.5841 | 0.6948 | 0.4996 | 0.6699 |
| **Final** | 0.0846 | 0.9015 | 0.0925 | 0.9413 | 0.8560 |

The final model performs binary classification 54.3500% more accurately than the baseline, and has 7.511 times less validation loss, 88.4107% greater sensitivity, and 27.7803% greater specificity. Although we treat overfitting as the most consequential metric, and the baseline overfits several times less than final model, the absolute values of overfitting for both models are relatively insignificant compared to the discrepancies mentioned earlier.

## 5.2    Notable Trends

All the Version 2 variations of ResNet (Resnet50V2, Resnet101V2, Resnet152V2) perform worse than their predecessors (Resnet50, Resnet101, Resnet152). Although all Version 2 variations suffer worse overfitting and classify less accurately, these models seem to minimize loss more effectively than their counterparts.

Surprisingly, although overfitting is primarily a disqualifying metric f or a model, many of the poorly- performing models had relatively low overfitting. Instead their results suffered due to abysmal performance for the following validation metrics: accuracy, loss, sensitivity, and specificity.

## 5.3 Implications for Deep Learning Research

Some of the novel methods presented in this paper can generate discussion regarding the controllable aspects of devising a deep neural network, such as the hyperparameters and network architecture. For instance, this work can encourage researchers to consider using architectures pre-trained on ImageNet images in their own image classification tasks. Moreover, the two-stage model selection framework can generalize to other applications of transfer learning and spur research for more unconventional modes of algorithm creation. We especially believe that the overall score heuristic and its dependency on non-parametric ranking can optimize hyper parameter tuning, or network structure design, such as neural architecture search (NAS) (Elsken et al., 2018).

## 5.4 Applications in Healthcare

The generalized healthy vs diseased format for our algorithm can offer a streamlined diagnosis for the general public, to help them simply determine whether to visit the doctor. We also think that the web-based format of the implementation expands the reach of the algorithm, as it can be used on any internet connected device. In order to enjoy widespread use, the predictions of the model should be confirmed through field trials in collaboration with medical experts as well as certified by the FDA and other regulatory agencies.

# 6 Future Model Training Improvements

## 6.1 Baseline Models and Hyperparameters

We should implement a more robust baseline model. In order to further validate the transfer learning approach, we can test the dataset with a Capsule Neural Network (Sabour et al., 2017). It has performed more favorably compared to conventional convolutional neural networks and has also exhibited compositional properties when evaluated on the MNIST dataset for handwritten digit recognition. The Capsule Net architecture could be also be used to evaluate spatio-temporal relationships between our model's features set and discover new correlations that may lead to improvements or deteriorations in ocular health.

Moreover, we can and should expand the scope of Stage 2, hyperparameter optimization. For instance, we can experiment with the number and type of additional layers added to the model architecture during Stage 1, or the epochs, batch size, and learning rate in Stage 2. To achieve this, we could use an open-source platform designed for machine learning model reproducibility, such as the CodaLab framework.

## 6.2    Improving Model through Web App: Active Learning

In the future, we could collect additional anonymized user input images not included in the original training data, with consent. Then, we can use approaches in unsupervised learning, where the labels to these images are unknown. Specifically, we could use a technique known as clustering, in which similar examples are grouped with one another in a graph. In this way, new data can be clustered with older training set data, from which our model can assign labels to previously unseen examples. As certain unlabeled images may not be similar to previous data distributions, we could also collaborate with physicians and medical professionals to get their domain expertise in classifying images that cross a heuristic uncertainty threshold.

## 6.3    Image Segmentation

The healthy set of images will have similar visual features, but because diseased images are combined into one category, their indicators can wildly vary. To combat the situation, we could employ image segmentation in conjunction with binary classification, to analyze biologically significant entities separately and model their relationships and attributes. For instance, a model could count the number of drusen particles to evaluate the severity of macular degeneration. Given a different example, the same model could identify an enlarged cup in the optic nerve, and assess the advancement of glaucoma.

## 6.4  Additional Improvements

Towards iteratively developing a better model, the most important progression we can make is in our data acquisition. We need to collect more data in general, and more importantly, a greater variety of diseases and imaging techniques must be represented for enhanced scalability.

An additional technique we can use in enlarging and improving the quality of our dataset is knowledge base construction. By representing data as a graph, we can extract implicit correlations from data as well as help standardize the format for large-scale data structure similar to a "Medical" ImageNet.